\documentclass[final]{cvpr}

\usepackage{times}
\usepackage{epsfig}
\usepackage{graphicx}
\usepackage{amsmath}
\usepackage{amssymb}

\usepackage[pagebackref=true,breaklinks=true,colorlinks,bookmarks=false]{hyperref}

\def\cvprPaperID{****} 
\def\confYear{CVPR 2021}

\begin{document}
	
	\title{Augmented 2D-TAN: A Two-stage Approach for Human-centric Spatio-Temporal Video Grounding}
	
	\author{Chaolei Tan\textsuperscript{1}\footnotemark[1], Zihang Lin\textsuperscript{1}\footnotemark[1], Jian-Fang Hu\textsuperscript{1}, Xiang Li\textsuperscript{2}, Wei-Shi Zheng\textsuperscript{1}\\
		\textsuperscript{1}Sun Yat-sen University, China\quad\textsuperscript{2}Meituan\\
		{\tt\small \{tanchlei, linzh59\}@mail2.sysu.edu.cn, hujf5@mail.sysu.edu.cn}\\
		{\tt\small lixiang82@meituan.com, wszheng@ieee.org}
	}

	\maketitle
	
    \renewcommand{\thefootnote}{\fnsymbol{footnote}}
    \footnotetext[1]{Equal contribution.} 
    \renewcommand{\thefootnote}{\arabic{footnote}}
    
	\begin{abstract}
		We propose an effective two-stage approach to tackle the problem of language-based Human-centric Spatio-Temporal Video Grounding (HC-STVG) task. In the first stage, we propose an Augmented 2D Temporal Adjacent Network (Augmented 2D-TAN) to temporally ground the target moment corresponding to the given description. Primarily, we improve the original 2D-TAN\cite{2dtan} from two aspects: First, a temporal context-aware Bi-LSTM Aggregation Module is developed to aggregate clip-level representations, replacing the original max-pooling. Second, we propose to employ Random Concatenation Augmentation (RCA) mechanism during the training phase. In the second stage, we use pretrined MDETR\cite{mdetr} model to generate per-frame bounding boxes via language query, and design a set of hand-crafted rules to select the best matching bounding box outputted by MDETR for each frame within the grounded moment. 
	\end{abstract}
	\section{Methodology}
	Human-centric spatio-temporal video grounding (HC-STVG) task \cite{hcstvg} aims to localize a spatio-temporal tube of the target person indicated by a language description. We tackle this problem with a two-stage approach which first localizes the temporal segment and then predicts the spatial location of the target person in each frame. In the first stage, we develop a temporal grounding model named Augmented 2D-TAN which predicts the temporal boundary (i.e., the start time and end time) of the target video segment corresponding to the given description. In the second stage, we generate per-frame proposals (bounding boxes) using MDETR\cite{mdetr} and find the best matching bounding box for each frame in the localized segment.
	\subsection{Augmented 2D-TAN for Temporal Localization}
\begin{figure}[t]
		\begin{center}
			\includegraphics[width=1\linewidth]{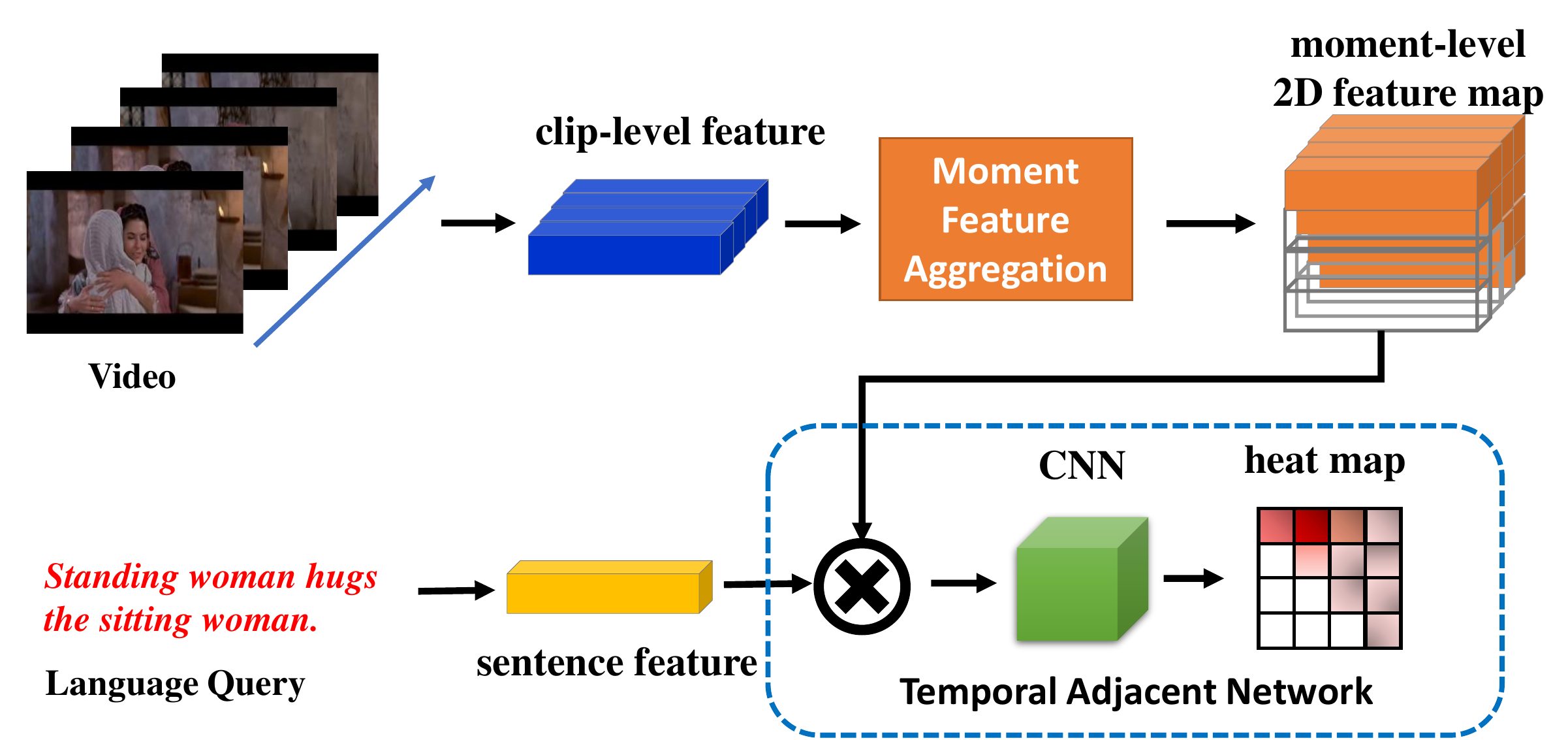}
		\end{center}
		\caption{Framework of Augmented 2D-TAN.}
		\label{fig:2D-TAN}
	\end{figure}

	We develop our Augmented 2D-TAN (as illustrated in Figure \ref{fig:2D-TAN}) grounding method based on 2D-TAN (2D Temporal Adjacent Networks)\cite{2dtan}. In this section, we will first briefly review 2D-TAN and its weaknesses, and then introduce how we improve it.
	
	\noindent\textbf{Review of 2D-TAN.}
	2D-TAN is a video grounding algorithm which first splits the input video into $N$ clips and extracts visual features\footnote{We use SlowFast Network\cite{slowfast} pretrained on Kinetics-600[\cite{kinetics-600} and AVA\cite{ava} to extract visual features.} for each clip. Then dense moment (a sequence of clips) proposals $\mathcal{P}$ are generated as $\mathcal{P} = \{P_{ij}=\left[C_i, C_{i+1}, ..., C_{j}\right] | 1\leq i\leq j\leq N\}$ where moment $P_{ij}$ contains the $i$-th clip $C_i$ to the $j$-th clip $C_j$. The feature of each moment proposal is extracted by a moment feature aggregation module $MFA(\cdot)$. Then a moment-level 2D feature map $M \in \mathbb{R}^{N\times N \times c}$ is constructed as:
	\begin{align}
	    M_{ij} = \begin{cases} MFA\left(\left[F_i, F_{i+1}, ... , F_j\right]\right) & i \leq j \\
	    \mathbf{0} & i > j
	                    \end{cases}
	\end{align}
	where $F_i$ is the feature of clip $C_i$. Then a Temporal Adjacent Network is employed to infer moment-wise matching score between $M$ and the sentence feature. The moment with the largest matching score is outputted as the grounding result. In the original implementation of 2D-TAN, $MFA(\cdot)$ is instantiated as the max-pooling operation which cannot model long-term temporal variation. Hence, the captured moment features are smoothed and the features of adjacent moments can be very similar, i.e., not discriminative. We propose a Bi-LSTM moment feature aggregation module to learn a more discriminative feature representation for each moment. And we further propose a Random Concatenation Augmentation strategy to augment the training data. We will introduce the Bi-LSTM module and the augmentation strategy in the following. 
	
	\begin{figure}[t]
		\begin{center}
			\includegraphics[width=0.9 \linewidth]{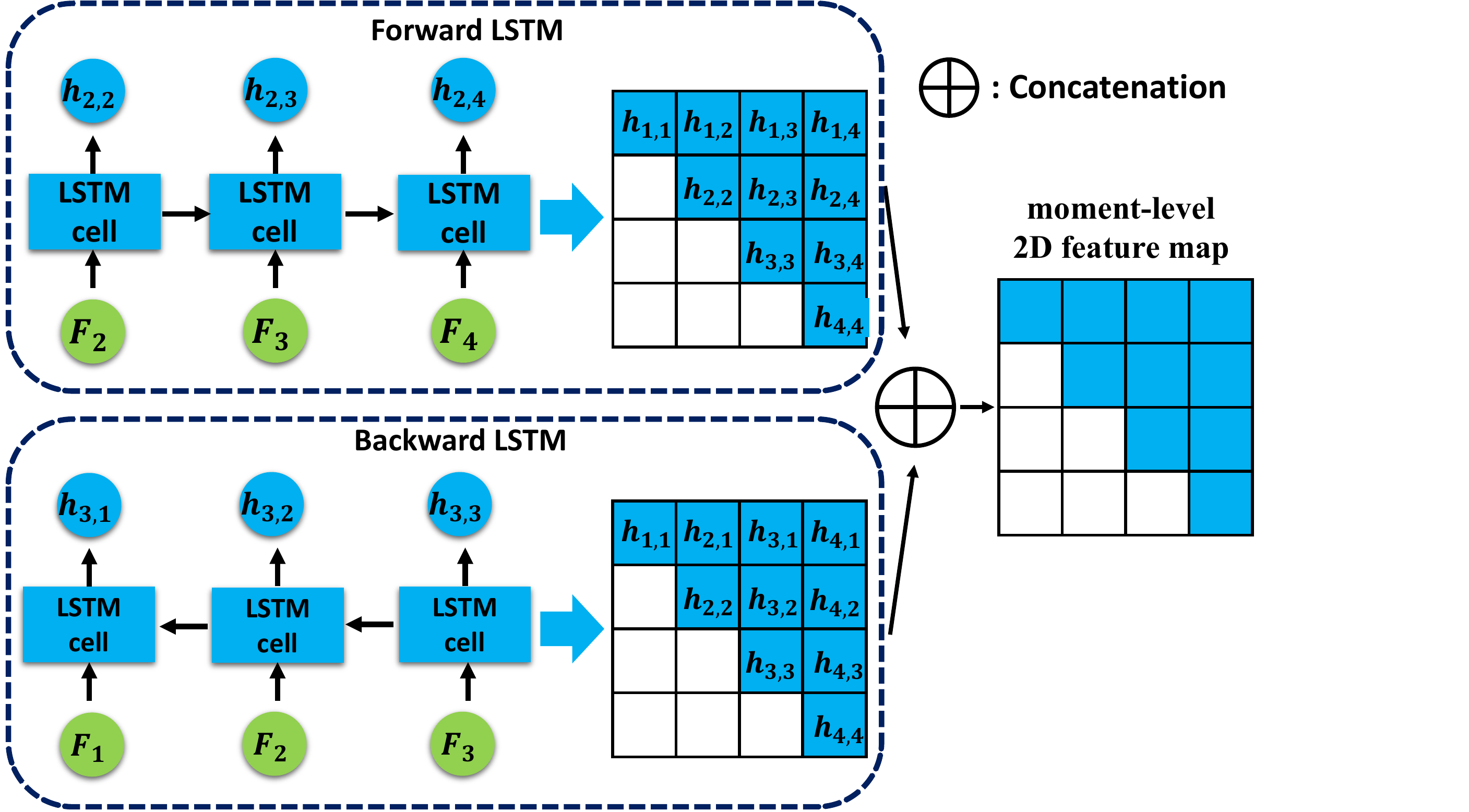}
		\end{center}
		\caption{The proposed Bi-LSTM aggregation module.}
		\label{fig:BiLSTM}
	\end{figure}
	
	\noindent\textbf{Bi-LSTM Aggregation Module.}
	As shown in Figure \ref{fig:BiLSTM}, the Bi-LSTM Aggregation Module consists of a forward LSTM and a backward LSTM. They are both effective for modeling the temporal variation in each moment. Thus, compared to the original 2D-TAN, the moment features learned with our Bi-LSTM aggegation module are more suitable for calculating the cross-modal matching score with the sentence feature. Note that the parameters of our Bi-LSTM are shared across all moments, so we can implement the inference in parallel to improve efficiency.

    \noindent\textbf{Random Concatenation Augmentation.}
	We observe severe overfitting in the original implementation of 2D-TAN and propose to randomly concatenate training samples to augment the training data. Specifically, we randomly select two videos (with similar frame rate) $V, V'$ and their corresponding language queries $Q, Q'$. We generate the augmented video as $V_{aug} = \{V_{(t)}\}_{t=\tau_s - \delta_1}^{\tau_e + \delta_2} || \{V'_{(t)}\}_{t=\tau_s' - \delta'_1}^{\tau_e' + \delta'_2}$, where $\tau_s$, $\tau_e$ indicate the start frame and end frame index of the ground truth moment, respectively. $\delta_1, \delta_2$ are random offsets and we ensure the number of frames in $V_{aug}$ is similar to that of $V$. $||$ represents concatenation operation along the temporal dimension. We randomly assign $Q$ or $Q'$ as the query of $V_{aug}$. During training, we perform this augmentation with a probability of 0.5 at each training step. The proposed augmentation strategy greatly enriches the training data and reduces the risk that the model simply learns to ground the salient segment. 

	\subsection{Per-frame Referring for Spatial Localization}
	We predict the spatial localization of the target person by a referring algorithm. Here, for each frame, we use the pretrained MDETR\cite{mdetr} model to predict a set of bounding boxes and their corresponding grounding text (several words extracted from the input sentence). Then we use a dependency parser\cite{parser} to find the subject of the input sentence. We filter out the bounding boxes (predicted by MDETR) whose grounding text doesn't contain the subject or contains more than one person. Finally, we select the bounding box with the longest grounding text from the remaining as our output. 
	
	


	\section{Experiments}

	As shown in in Table \ref{tab:ablation}, the proposed Random Concatenation Augmentation (RCA) and Bi-LSTM aggregation module both bring consistent performance improvement, which demonstrates their effectiveness. To further boost the performance, we ensemble 10 models for Augmented 2D-TAN and achieve 31.9\% viou on the test set. The ensembled version achieves the best performance on HC-STVG test set\cite{hcstvg} compared to other methods\footnote{Results are copied from \url{http://www.picdataset.com/challenge/leaderboard/hcvg2021}} as shown in Table \ref{tab:HCVG}.
	\begin{table}
		\small
		\begin{center}
			\begin{tabular}{|l|c|c|c|c|c|}
				\hline
				Method & viou@0.3 & viou@0.5 & tiou & viou & split\\
				\hline\hline
				2D-TAN & 47.2 & 16.7 & 53.4 & 29.5 &val \\
				+RCA & 48.8 & 16.7 & 54.1 & 30.0 & val \\
				+Bi-LSTM & \textbf{50.4} & \textbf{18.8} & \textbf{55.2}  & \textbf{30.4} &val \\
				\hline \hline
				2D-TAN$^{\dagger}$ & 47.7 & 20.2 & 53.3 & 29.4 & test\\
				Ours$^{\dagger}$ & \textbf{52.7} & \textbf{22.2} & \textbf{56.5} & \textbf{31.9} &test\\
				\hline
			\end{tabular}
		\end{center}
		\caption{Ablation study on HC-STVG dataset\cite{hcstvg}. $\dagger$: Ensemble.}
		\label{tab:ablation}
	\end{table}
	
	\begin{table}
		\small
		\begin{center}
			\begin{tabular}{|l|c|c|c|c|}
				\hline
				Method & viou@0.3 & viou@0.5 & tiou & viou\\
				\hline\hline
				easy\_baseline2.5 & 48.6 & 24.4 & \textbf{66.0} & 30.9 \\
				try10 & 47.9 & \textbf{27.0} & 53.8 & 31.3 \\
				2stage & 50.2 & 22.5 & 55.9 & 31.4 \\
				\hline \hline
				Ours & \textbf{52.7} & 22.2 & 56.5 & \textbf{31.9} \\
				\hline
			\end{tabular}
		\end{center}
		\caption{Comparison results on HC-STVG test set\cite{hcstvg}.}
		\label{tab:HCVG}
	\end{table}
	
	\section{Conclusion}
	In this report, we introduce our proposed Augmented 2D Temporal Adjacent Network (Augmented 2D-TAN) for HC-STVG task. In virtue of the proposed temporal context-aware Bi-LSTM Aggregation Module and Random Concatenation Augmentation Mechanism, Augmented 2D-TAN is powerful to capture moment features and less prone to overfit the training data. Our proposed method has achieved the state-of-the-art performance on HC-STVG test set\cite{hcstvg}.

	{\small
		\bibliographystyle{ieee_fullname}
		\bibliography{main}
	}
	
\end{document}